\documentclass[fleqn,10pt]{wlscirep}
\usepackage[utf8]{inputenc}
\usepackage[T1]{fontenc}
\usepackage{lineno}

\usepackage{subfig}
\usepackage{graphicx}
\usepackage{amsmath}
\title{An Experimental Study of Data Heterogeneity in Federated Learning Methods for Medical Imaging}

\author[1]{Liangqiong Qu}
\author[1]{Niranjan Balachandar}
\author[2]{Daniel L Rubin}
\affil[1]{Department of Biomedical Data Science at Stanford University, Stanford, CA 94305,
USA}
\affil[2]{Department of Biomedical Data Science and Department of Radiology at Stanford
University, Stanford, CA 94305, USA.}

\begin{abstract}
Federated learning enables multiple institutions to collaboratively train machine learning models on their local data in a privacy-preserving way.  However, its distributed nature often leads to significant heterogeneity in data distributions across institutions. In this paper, we investigate the deleterious impact of a taxonomy of data heterogeneity regimes on federated learning methods, including quantity skew, label distribution skew, and imaging acquisition skew. We show that the performance degrades with the increasing degrees of data heterogeneity. We present several mitigation strategies to overcome performance drops from data heterogeneity, including weighted average for data quantity skew, weighted loss and batch normalization averaging for label distribution skew.
The proposed optimizations to federated learning methods improve their capability of handling heterogeneity across institutions, which provides valuable guidance for the deployment of federated learning in real clinical applications.
\end{abstract}
\begin{document}

\flushbottom
\maketitle

\thispagestyle{empty}

\section{Introduction}
Deep learning techniques have demonstrated state-of-the-art performances in a wide range of computer vision and automatic clinical tasks, such as classification of natural images, detection and diagnosis of cancer, and clinical prediction \cite{krizhevsky2012imagenet, coudray2018classification, qu2020synthesized,mobadersany2018predicting,wang2020deep}.
However, the advancement of deep learning techniques is heavily dependent on the amount and diversity of the data in the training dataset.  Many deep learning models are currently trained using data from few centers and generally do not perform well in new data or in clinical practice. Cohort sizes at single institution or even in public data repositories such as The Cancer Imaging Archive (TCIA) are often small, especially for rarer diseases or for certain patient populations (e.g., molecular variants in gliomas) \cite{clark2013cancer, qu2019wavelet, kaushal2020geographic}. Aggregating data from multiple institutions is not always feasible due to regulatory, technical and patient privacy concerns.
Federated learning, where computations are performed locally at each institution without sharing data, is promising for accessing large, representative data to train robust deep learning models that have greater generalizability and lower risk of model bias.

Numerous federated learning methods have emerged in past decades \cite{su2015experiments,mcmahan2016communication,lin2017deep,chang2018distributed,vepakomma2018split}, such as Federated stochastic gradient descent (FedSGD) \cite{su2015experiments,mcmahan2016communication}, Federated averaging algorithm (FedAVG) \cite{mcmahan2016communication}, and Cyclical weight transfer (CWT) \cite{chang2018distributed}. Despite the promising progress, existing methods generally do not account for the presence of data heterogeneity among institutions.
Most federated learning methods usually assume independent and identically distributed (IID) data among institutions, which is unlikely to hold in real federated learning settings in healthcare. Few recent studies have explored the performance of federated learning methods on non-IID data partitions that have label distribution skew \cite{hsu2019measuring,hsieh2019non,Niranjan2020}. For example, Hsieh \emph{et al.} \cite{hsieh2019non} conducted a series of experimental studies to show the impact of label distribution skew on federated learning methods. However, in addition to label distribution skew, the data at different institutions are usually heterogeneous in other ways, such as data quantity skew and imaging acquisition skew. For example, large academic university hospitals generally have substantially larger datasets than small community hospitals, and they differ in equipment vendors and imaging equipment parameters.

In this paper, we present an experimental study on two medical image classification tasks, to investigate the impact of a taxonomy of data heterogeneity regimes on federated learning methods, including quantity skew, label distribution skew, and imaging acquisition skew (e.g., different hospitals may use different imaging equipment vendors and acquisition protocols). Contributions of this paper are summarized as follows.

1) We present the first thorough research to study the impact of a taxonomy of data
heterogeneity regimes on several widely used federated learning methods with medical image data. Our study provides valuable guidance for the deployment of federated learning in real clinical applications. %, and also provides directions for future optimization study of federated learning methods.

2) We show that the performance of the federated learning methods in our study degrades with the increasing degrees of data heterogeneity, and
the rate of decrease in performance is determined by the degree of deviation from homogenous distributions. %of data heterogeneity.

3) We propose a variety of optimization strategies to mitigate the performance loss for quantity skew and label distribution skew,
including weighted average strategy for data quantity skew, and weighted loss strategy for label distribution skew.

4) We study the influence of the Batch Normalization (BN) for FedAVG, we show that averaging the mean and variance of BN across institutions during FedAVG training is a simple and flexible alternative to mitigate skew-induced performance loss of BN.

%We then investigate two model selection schemes for FedAVG according to the real demand of federated learning:
%the standard model selection from central server with averaged BN layer for the demand of a global generalized model works for multiple-institutions,
%and model selection with one epoch local finetuning without average BN layer for the demand of a specialized model works better on local institutions.

\section{Study Materials and Experimental Setup}
\subsection{Study Methods}
We employed three popular federated learning methods in our study: two parallel federated learning methods (FedSGD \cite{su2015experiments,mcmahan2016communication} and FedAVG \cite{mcmahan2016communication}), and CWT \cite{chang2018distributed}.
We used the model trained with the centrally hosted data as a baseline method, termed as centrally hosted.
\iffalse
\begin{itemize}
\item \textbf{FedSGD \cite{su2015experiments,mcmahan2016communication}}, involves frequent transferring of gradients from individual institutions to a central server, computing weight updates using institutional gradients at the central server, and transferring updated weights back to individual institutions.
\item \textbf{FedAVG \cite{mcmahan2016communication}}, similar to FedSGD, FedAVG involves 1) frequent transferring of weights from individual institutions to a central parameter server, 2) averaging weight computation at the central server, and 3) transferring averaged weights back to individual institutions.
\item \textbf{CWT \cite{chang2018distributed}}, involves training for a fixed number of iterations at one institution, and cyclically transferring weights to the next training institution until model convergence.
\end{itemize}
\fi

\textbf{FedSGD \cite{su2015experiments,mcmahan2016communication}}, involves frequent transferring of gradients from individual institutions to a central server, computing weight updates using institutional gradients at the central server, and transferring updated weights back to individual institutions.

\textbf{FedAVG \cite{mcmahan2016communication}}, involves frequent transferring of weights from individual institutions to a central parameter server, averaging weight computation at the central server, and transferring averaged weights back to individual institutions.

\textbf{CWT \cite{chang2018distributed}}, involves training for a fixed number of iterations at one institution, and cyclically transferring weights to the next training institution until model convergence.

\subsection{Dataset}
We evaluated on Alzheimer’s Disease Neuroimaging Initiative (ADNI) Dataset \footnote{\url{http://adni.loni.usc.edu/}} and Diabetic Retinopathy (Retina) Dataset \cite{KaggleRetina}:

\textbf{ADNI Dataset} provides a longitudinal multi-institutional observation study on Alzheimer's disease patients, mild cognitive impairment subjects, and healthy elders controls. %It acquires Magnetic resonance imaging (MRI), including MRI and PET images, genetics, cognitive tests, CSF and blood biomarkers as predictors of the disease urine serum, and cerebrospinal fluid (CSF) biomarkers, as well as clinical/psychometric assessments
      For simplicity, we only use PET scans from (18F)-fluorode-oxyglucose PET to predict the summary standard uptake value ratios (SUVRs) status. Specifically, we discretized the continuous SUVRs into 4 classes, 2 classes of equal size for SUVR values below the 1.11 cutoff \cite{landau2012amyloid}, and 2 classes for the SUVR values above. We utilized a total of 2603 florbetapir PET scans (from 1235 subjects) in our study. We randomly divided the scans into 3 subsets: 1896 scans for training, 314 scans for validation, and the remaining for testing. The scans for the same subject were divided into same subset. %The samples were acquired from 62 different sites by 40 different types of scanners.

\textbf{Retina Dataset} \cite{KaggleRetina} consists of 17,563 pairs of right and left color digital retinal fundus images. Each image provided by this dataset includes a rating on a scale of 0 to 4 according to the presence of diabetic retinopathy. Following the setting in \cite{chang2018distributed}, the labels were binarized to Healthy (scale 0) and Diseased (scale 2, 3 or 4) in our study, while the mild diabetic retinopathy images (scale 1) were excluded. Additionally, only left eye images were used to remove the confusion from using multiple images for the same patient. We utilized a total of 6000 subjects for training, 3000 for validation, and 3000 for testing.

\subsection{Experimental Setup}
We used a full communication setting for FedSGD, i.e., update the gradients between individual institutions and central server every iteration.
For FedAVG, we set the number of training passes on each local institution to $Q_i/B$, where ${Q_i}$ is the quantity of training samples in local institution $I_i$ and $B$ is the local minibatch size. CWT involves training the same fixed number of iterations on each institutions. Here we set the number of iterations to $Q/(B\times n)$, where $Q$ is the quantity of training samples of centrally hosted data, $n$ is the number of institutions involving in federated learning.

We applied ResNet34 \cite{he2016deep} as our baseline DNN architecture. All the methods were implemented in Pytorch and optimized with SGD. The training parameters (such as learning rate and minibatch size) were tuned to make sure the baseline centrally hosted achieved best performance. For a fair comparison, we then used these training parameters in all the federated learning methods we compared. All methods were trained with enough epochs until model convergence.

\section{Experiments and Results}
In this section, we study the impact of a taxonomy of data heterogeneity regimes on federated learning methods, including quantity skew, label distribution skew
and imaging acquisition skew. We then present several mitigation strategies to overcome performance drops from data heterogeneity.
\subsection{Quantity Skew} \label{sec_quantity_skew}
\begin{figure*}
\scriptsize
	\begin{center}
		\begin{tabular}{cc}
\vspace{1mm}
 \includegraphics[width=0.47\linewidth]{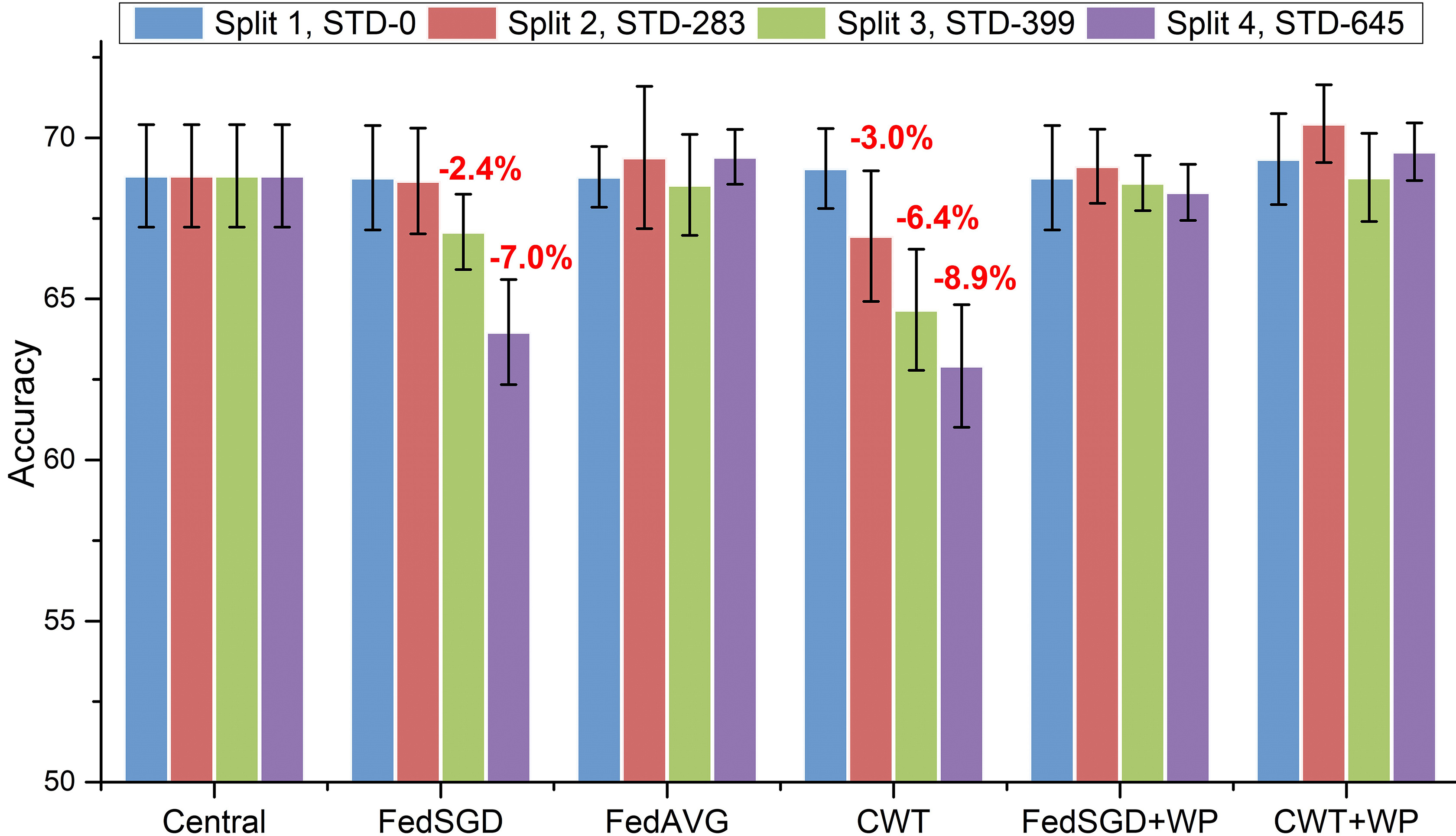}&
  \includegraphics[width=0.45\linewidth]{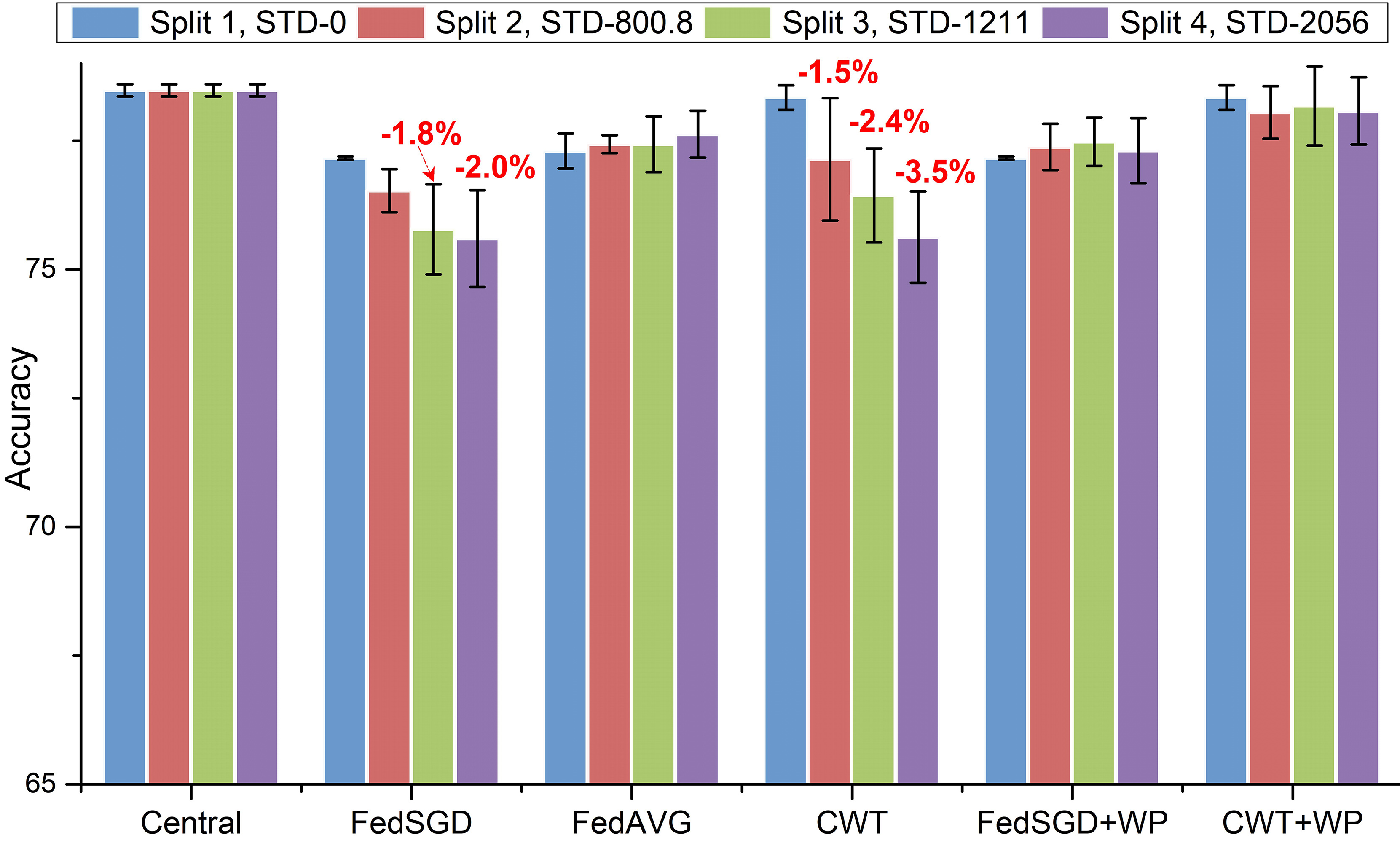}\\
   ADNI & Retina  \\
		\end{tabular}
	\end{center}
	\caption{Test accuracy on data partitions with quantity skew\protect\footnotemark[2]. The performance drop rate of data partitions with quantity skew from homogenous data split 1 is shown when it is larger than 1\%. }
	\label{fig:Performance_size_var}
\end{figure*}
Quantity skew is one common way that causes data to deviate from a homogenous distribution. Large academic university hospitals generally have substantially larger datasets than small community hospitals. We created 4 sets of data partitions with variable sample sizes across simulated institutions to study the impact of quantity skew on federated learning methods (see supplementary file for detailed data partitions). Each data partition consists of 4 stimulated institutions and each institution shares the same feature distribution and label distribution.
We use sample standard deviation (STD) of the sample size across institutions to measure the degree of quantity skew.
\footnotetext[2]{Mean and standard deviation test accuracies were obtained with 4 runs. We use the same setting for the following experiments.}
\iffalse
\begin{table}
\renewcommand{\tabcolsep}{0.7pt}
\footnotesize
  \centering
  \caption{Synthetic data partitions with quantity skew. STD is sample standard deviation of the training sample size across institutions.}
  \label{table:size_partition}
  \subfloat[Partitions on ADNI]{%
    \hspace{.5cm}%
    \begin{tabular}{|c|c|c|c|c|c|}
        \hline
        Splits       & Inst1 & Inst2  & Inst3  & Inst4 & STD \\
        \hline
        \hline
        Split 1     & 474       &  474 &  474 &  474 & {0}  \\
        Split 2     & 299       &  317 &  385 &  895  & {283.1} \\
        Split 3     & 113       &  211 &  579 &  993  & {399.9} \\
        Split 4     & 66        & 111  &  282 &  1437 & {648.7} \\

        \hline
    \end{tabular}%
    \hspace{.5cm}%
  }
  \subfloat[Partitions on Retina]{%
    \hspace{.5cm}%
    \begin{tabular}{|c|c|c|c|c|c|}
        \hline
        Splits      & Inst1 & Inst2  & Inst2  & Inst2 & STD \\
        \hline
        \hline
        Split 1     & 1500       &  1500 &  1500 &  1500 & {0}  \\
        Split 2     & 750       &  960 &  1800& 2490 & {800.8} \\
        Split 3     &315	& 850	& 1750	& 3085	& {1211} \\
        Split 4     &208	& 350	& 889	& 4553	& {2056} \\
        \hline
    \end{tabular}%
   % \hspace{.5cm}%
  }
\end{table}
\fi

Fig.~\ref{fig:Performance_size_var} shows that all the federated learning methods achieve comparable performance to the centrally hosted baseline in homogenous data split 1. However, the performance of FedSGD and CWT degrade with the increasing degree of skew, e.g, 7.0\% and 8.9\% drop rates of FedSGD and CWT on ADNI dataset with split 4.
All the institutions at FedSGD contributed equally on the gradient update at each training iteration, ignoring the impact of quantity skew. Different from FedSGD, FedAVG averages the model weights after one epoch train on the whole local institutional dataset, without reusing the small dataset for weights update, thus works well on quantity skew. We then introduced a weighted average strategy for FedSGD, termed as FedSGD+WP, where gradients from each institutions were not treated equally but proportional to the institutional training sample size. Similarly, we applied proportional training sample sizes \cite{Niranjan2020} to CWT for addressing quantity skew, termed as CWT+WP. Shown in Fig.~\ref{fig:Performance_size_var}, with the proposed weighted average strategy and proportional training sample sizes strategy, both FedSGD and CWT achieve promising performance on data partitions with quantity skew.

\subsection{Label Distribution Skew} \label{sec_label_skew}
Our study on quantity skew assumes homogenous label distribution across institutions, which is not always true in real applications.
In fact, label distribution may vary across institutions even when they share the same label annotations. For example, lupus is much more common in Black, Asian people than White people.
\iffalse
\begin{figure}
\centering
\includegraphics[width=0.95\linewidth]{picture/data_statis_label_var_ADNI.png}
\caption{Data partitions on ADNI dataset with label distribution skew. Large Kolmogorov-Smirnov (KS) value indicates higher degree of label distribution skew.}
	\label{fig:Data_statis_label_var_ADNI}
\end{figure}
\fi
We created 4 sets of data partitions with label distribution skew (same data quantity) by controlling the fraction of non-IID data. We used the mean Kolmogorov-Smirnov (KS) statistic between every two institutions to measure the degree of label distribution skew. Specifically, KS=0 indicates homogenous label distributions, and 1 indicates totally different label distributions across institutions. See supplementary file for detailed data partitions and its corresponding KS value. % is shown in Fig.~\ref{fig:Data_statis_label_var_ADNI}.

\begin{figure*}
\scriptsize
	\begin{center}
		\begin{tabular}{cc}
 \includegraphics[width=0.5\linewidth]{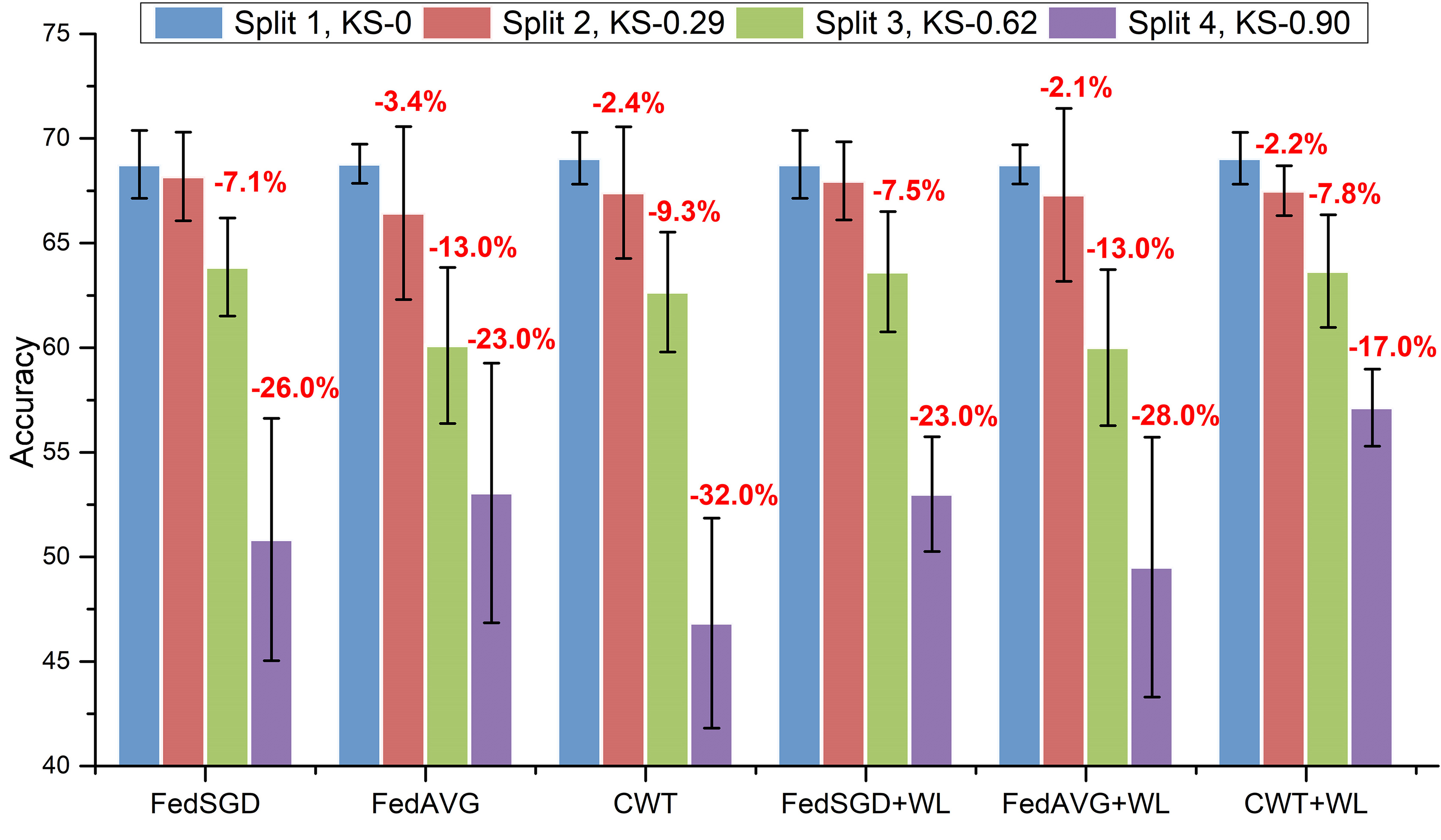}&
  \includegraphics[width=0.5\linewidth]{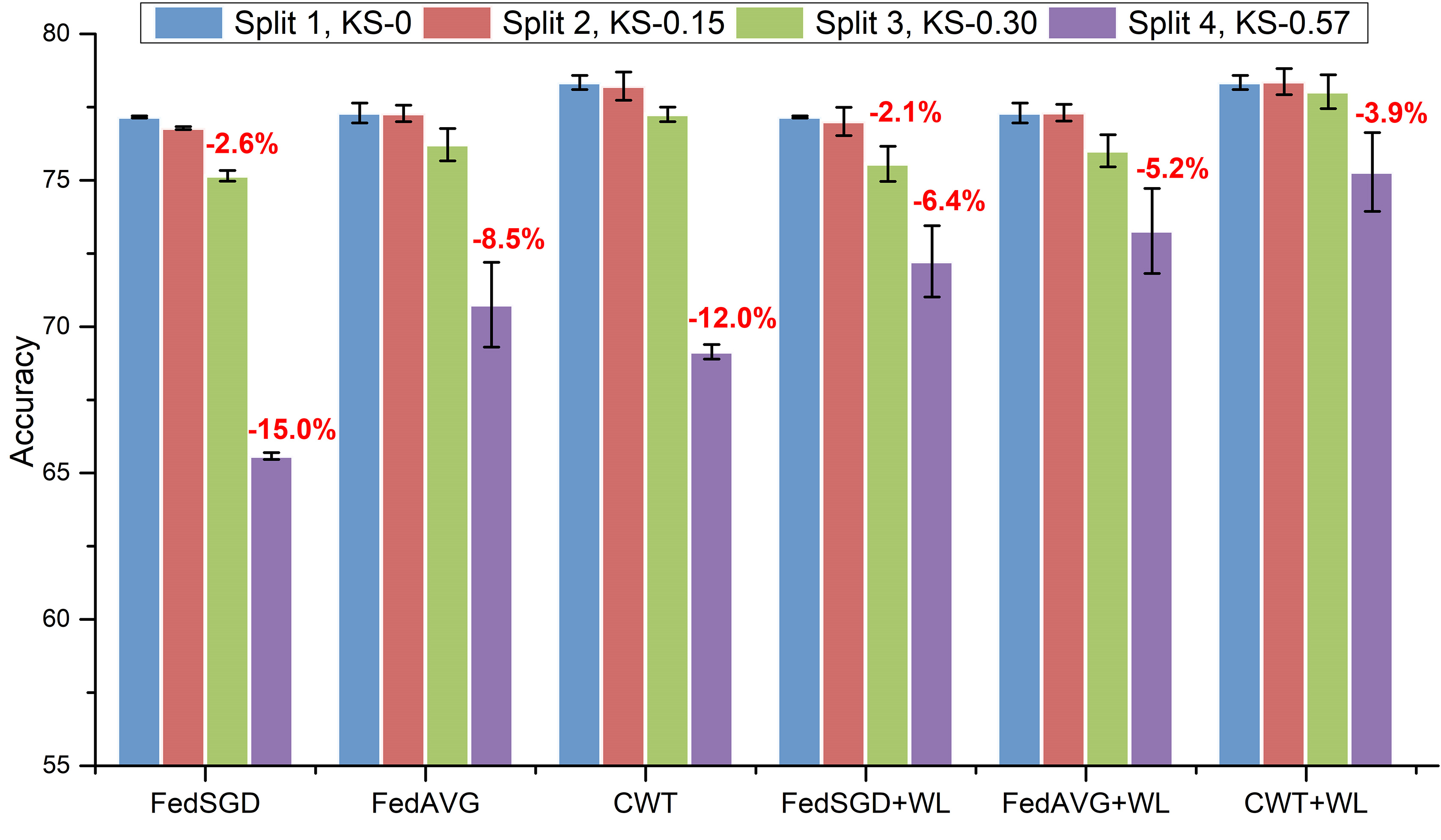}\\
   ADNI & Retina  \\
		\end{tabular}
	\end{center}
	\caption{Test accuracy on data partitions with label distribution skew. The performance drop rate of data partitions with label distribution skew from homogenous data split 1 is shown when it is larger than 1\%. }
	\label{fig:Performance_label_var}
\end{figure*}

Fig.~\ref{fig:Performance_label_var} shows that all the compared federated learning methods are vulnerable to label distribution skew. The performance drop rates on split 4 (with KS = 0.90) of ADNI dataset reaches to 26.0\%, 23.0\%, and 32.0\% of FedSGD, FedAVG, and CWT, respectively.
%Interestingly, even the FedSGD, with frequent communication of gradient update every iteration, also cannot retain the model quality in data partitions with high label distribution skew.
One common approach for addressing class imbalance in standard DNNs is to introduce weighted factors into loss function. Similarly, we also applied a class weighted cross-entropy loss (WL) to tackle the label distribution skew in federated learning methods:
\begin{equation}\label{weighted_CW}
 \text{WL} = - \Sigma_{j=1}^c \alpha_j y_{x,j} \log(p_{x,j}),
\end{equation}
where $C$ is the number of categories,  $y_{x,j}$ is the ground-truth binary indicator with 1 when class label $j$ is the correct category for sample $x$ and 0 else, and $p_{x, j}$ is the model prediction probability that sample $x$ has class label $j$,
$\alpha_j$ is the weighted factor for class label $j$ and is defined as:

\begin{equation}\label{factors}
  \alpha_j = \frac{\mathop {\max }\{n_1, n_2, ..., n_C\}}{n_j},
\end{equation}
where $n_j$ indicates the number of samples with class label $j$. WL is applied on both the training stage and the cross-validation model selection stage for better convergence.

WL mitigates the performance drops on data partitions with certain degrees of label distribution skew (see Fig.~\ref{fig:Performance_label_var}). For example, the performance drop rate of CWT+WL is increased from 32.0\%, and 12.0\% to 17.0\% and 3.9\% on split 4 of ADNI and Retina dataset, respectively. However, the improvement on parallel federated learning (FedSGD and FedAVG) with large degree of skew is limited. FedAVG+WL even works worse than FedAVG on split 4 of ADNI dataset, e.g., a 28.0\% drop rate compared to original 23\%. Note that WL has also been described in \cite{Niranjan2020}, but they only applied them to CWT for 2-label classification tasks. Here, we extend it to general federated learning approaches and multi-label classification tasks.

\begin{figure*}[h]

\scriptsize
	\begin{center}
		\begin{tabular}{ccc}
\includegraphics[width=0.33\linewidth, height =0.25\linewidth]{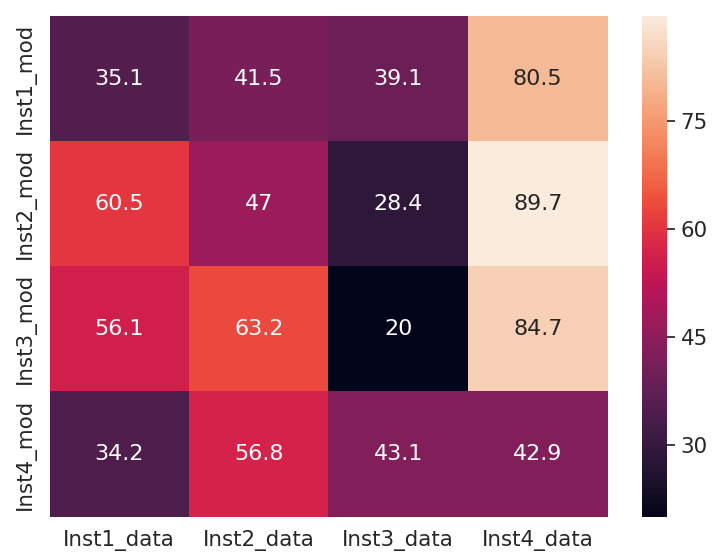}&
\includegraphics[width=0.33\linewidth, height =0.25\linewidth]{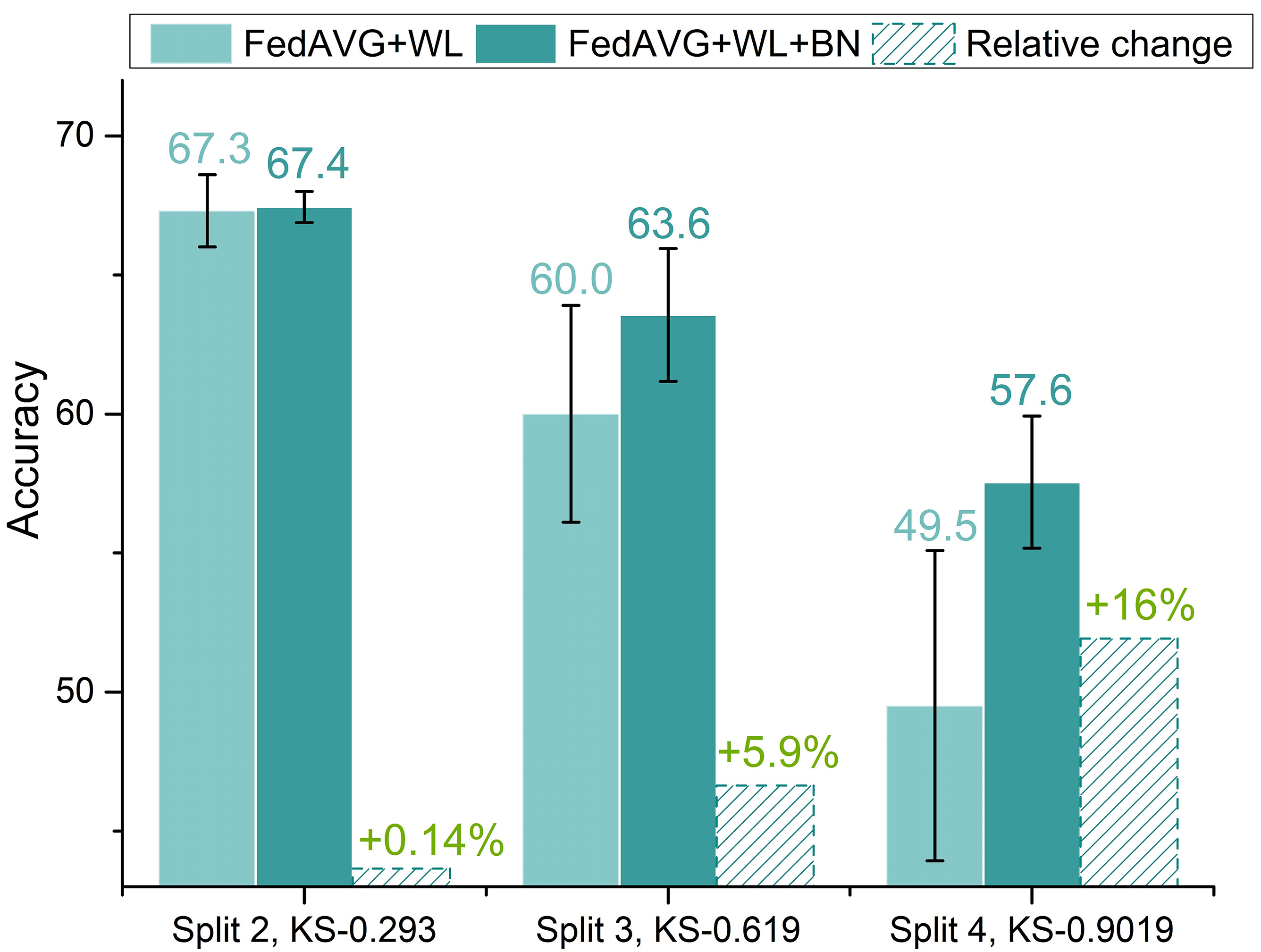}&
  \includegraphics[width=0.33\linewidth, height =0.25\linewidth]{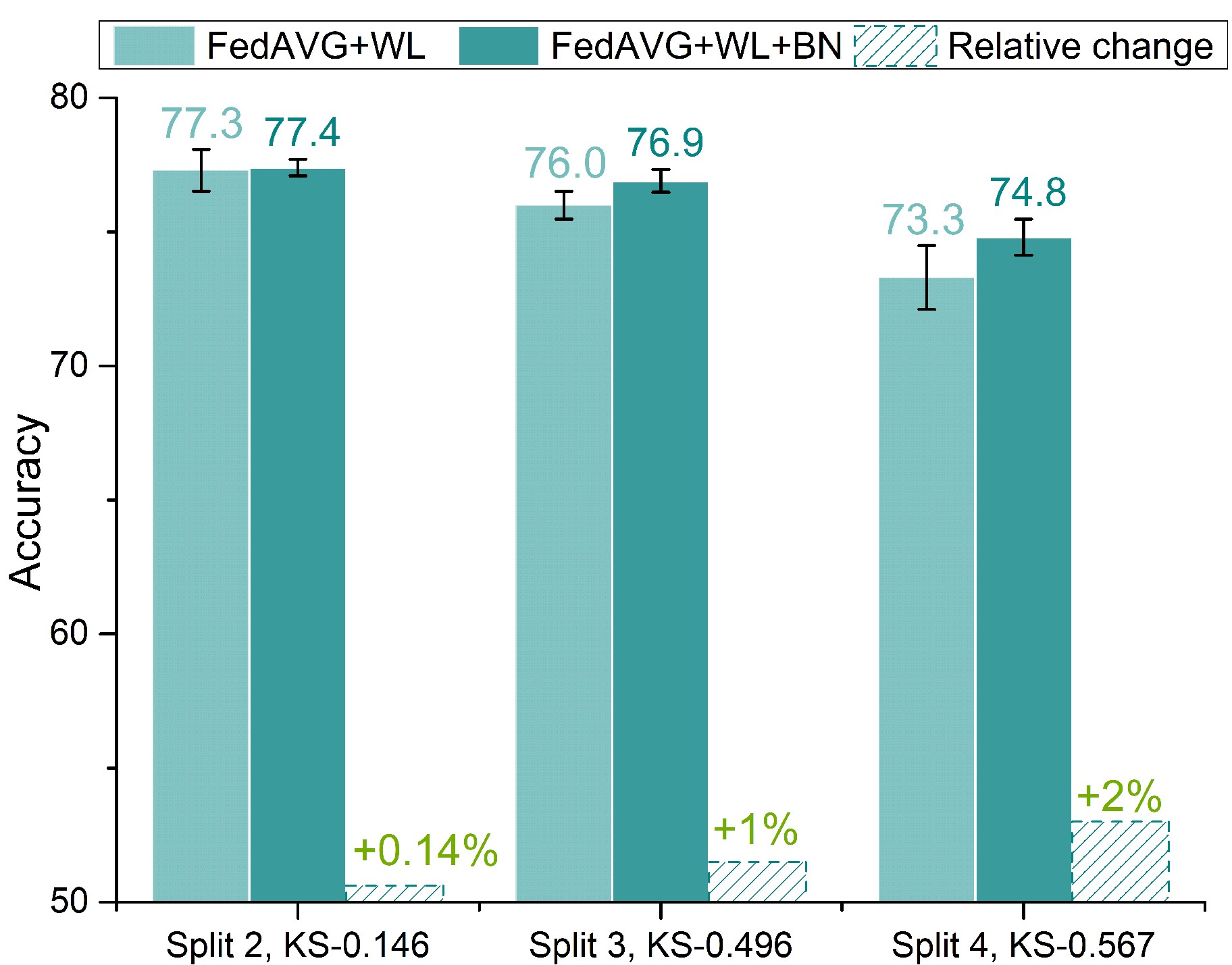}\\
(a) Split 4 of ADNI &  (b) ADNI & (c)Retina\\
		\end{tabular}
	\end{center}
	\caption{(a): Prediction accuracy of each institutional model on each institutional test dataset of split 4 on ADNI datasest.  (b) and (c): Performance analysis with different averaging settings of BN in FedAVG on ADNI and Retina dataset, respectively. FedAVG+WL+BN helps mitigate the skew-induced accuracy loss of BN by averaging the estimated mean and variance of BN.}
	\label{fig:FedAVG_BN}
\end{figure*}

To investigate the worse performance of FedAVG+WL, we show the prediction accuracy of each institutional model on each institutional test dataset\footnotemark[3]\footnotetext[3]{Same label distribution for local institutional testing and training dataset.} (see Fig.~\ref{fig:FedAVG_BN}(a)).
It is worth noting that the institutional model performs even worse on its own testing dataset than the models from other institutions, e.g., worse diagonal performance of row 1 and row 3 in Fig.~\ref{fig:FedAVG_BN}(a). %This indicates that the model maybe not applicable to its own test dataset after directly model average on central

This then raises the question: why the institutional models after model average differ a lot?  Actually, the models across institutions only differ in the BN layer after model averaging on central server. Thus, BN is the main factor that causes the discrepancy between models across institutions. BN is a widely used technique aiming to stabilize and accelerate the DNN training. It normalizes each layer's inputs with minibatch mean and variance during training, and an estimated global mean and variance is used during testing.
%Since it is impractical to attain the global mean and variance information, in real implementation, the minibatch mean and variance is applied during training.
In standard implementation, FedAVG and FedSGD do not average the estimated mean and variance of each layer, which then causes the discrepancy between models across institutions. To this end, we modified the averaging setting of BN, also averaged the estimated mean and variance across institutions during federal training. Fig.~\ref{fig:FedAVG_BN} shows results with averaged BN settings. FedAVG+WL+BN generates superior performance than the standard FedAVG+WL setting, e.g., a 16.0\% percentage increase is shown on split 4 of ADNI dataset. \cite{hsieh2019non} also indicates that DNNs with BN is vulnerable to label distribution skew, they show that group normalization \cite{wu2018group} avoids the skew-induced accuracy loss of BN. Our experiments are complementary to theirs and provide a simple and flexible alternative to help mitigate the skew-induced performance loss of BN in federated learning settings.

\subsection{Imaging Acquisition Skew}
In medical domain, there has been a longstanding debate about applying medical imaging standardization throughout the imaging industry \cite{gillies2016radiomics,lambin2017radiomics}. Modern X-Ray, MR imaging, and PET units allow for wide variations in imaging acquisition settings, which may result in significant differences in the generated images across institutions, even for the same underlying disease. %A large number of recent research also show that a well-trained deep model has poor generalizability when applied to data from different institutions.

In this section, we show the impact of imaging acquisition skew on federated learning methods with two sets of experiments.
We first simulated three sets of data partitions (split 1 - split 3), and then generated a real data partition on ADNI dataset according to scanner vendors (split 4).
\iffalse
\begin{itemize}
  \item \textbf{Split 1} (Simulated IID partition): Same as split 1 in Sec.~\ref{sec_quantity_skew} and Sec.~\ref{sec_label_skew}.
  \item \textbf{Split 2} (Simulated partition with resolution skew): Decreasing image resolution in Institution 1 to 4 of split 1 with a factor of 4, 3, 2, 1, respectively.
  \item \textbf{Split 3} (Simulated partition with signal-to-noise-ratio skew): Degrading images in split 1 with various types of noises and blurring, such as Gaussian, Speckle, Poisson noise and motion blur. Institution 1: Gaussian noise, Institution 2: motion blur, Institution 3: mixture of
      noise/blurring but dominated by gaussian noise, Institution 4: mixture of noise/blurring but dominated by motion blur. Examples of synthetic gaussian noise and motion blur are shown in Fig.~\ref{fig:synthetic_img_var}.
  \item \textbf{Split 4} (Real partition according to scanner vendors): ADNI dataset were acquired from 7 manufacturers, such as GE Healthcare, Philips Healthcare, Siemens Healthcare and etc.  We resplit the dataset into 4 institutions according to the scanner vendors. Least quantity skew and label distribution skew across institutions are ensured in this new split. %An example of images captured with different scanner vendors were shown in Fig~\ref{fig:scanner_vendors},
      %they differs in either resolution, contrast, or intensity distributions.
\end{itemize}
\fi

\textbf{Split 1} (Simulated IID partition): Same as split 1 in Sec.~\ref{sec_quantity_skew} and Sec.~\ref{sec_label_skew}.

\textbf{Split 2} (Simulated partition with resolution skew): Decreasing image resolution in Institution 1 to 4 of split 1 with a factor of 4, 3, 2, 1, respectively.

\begin{figure*}
\scriptsize
	\begin{center}
		\begin{tabular}{cccc}
\includegraphics[width=0.215\linewidth, height=0.225\linewidth]{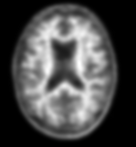}&
\includegraphics[width=0.215\linewidth, height=0.225\linewidth]{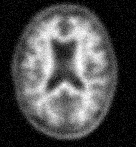}&
\includegraphics[width=0.23\linewidth, height=0.225\linewidth]{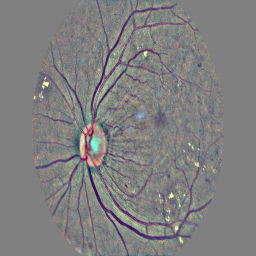}&
  \includegraphics[width=0.23\linewidth, height=0.225\linewidth]{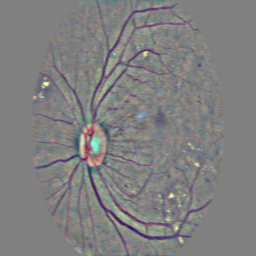}\\
  Orig img.& Orig img. + Gaussian & Orig img. & Orig. img + motion blur\\
		\end{tabular}
	\end{center}
	\caption{Examples of images with synthetic gaussian noise and motion blur. }
	\label{fig:synthetic_img_var}
\end{figure*}

\textbf{Split 3} (Simulated partition with signal-to-noise-ratio (SNR) skew): Degrading images in split 1 with various types of noises and blurring, such as Gaussian, Speckle, Poisson noise and motion blur. Institution 1: Gaussian noise, Institution 2: motion blur, Institution 3: mixture of
      noise/blurring but dominated by gaussian noise, Institution 4: mixture of noise/blurring but dominated by motion blur. Examples of synthetic gaussian noise and motion blur are shown in Fig.~\ref{fig:synthetic_img_var}.

\textbf{Split 4} (Real partition according to scanner vendors): ADNI dataset were acquired from 7 manufacturers, such as GE Healthcare, Philips Healthcare, Siemens Healthcare and etc.  We resplit the dataset into 4 institutions according to the scanner vendors. Least quantity skew and label distribution skew across institutions are ensured in this new split. %An example of images captured with different scanner vendors were shown in Fig~\ref{fig:scanner_vendors},

%Note that all the deterioration procedure were directly performed on the post-processing images. For example, the final resolution in Retina dataset is $64 \times 64$, $85 \times 85$, $128 \times 128$, and $256 \times 256 $ in Split 2.
\iffalse
\begin{figure*}
\scriptsize
	\begin{center}
		\begin{tabular}{ccccc}
  \includegraphics[width=0.192\linewidth]{picture/I212892.png}&
\includegraphics[width=0.192\linewidth]{picture/I411272.png}&
\includegraphics[width=0.192\linewidth]{picture/I195573.png}&
  \includegraphics[width=0.192\linewidth]{picture/I337063.png}&
   \includegraphics[width=0.192\linewidth]{picture/I711382.png} \\
   Philips & Siemens & CPS & GEMS & MiE\\
		\end{tabular}
	\end{center}
	\caption{An example of PET cases captured with different scanner vendors. }
	\label{fig:scanner_vendors}
\end{figure*}
\fi

\begin{figure*}
\scriptsize
	\begin{center}
		\begin{tabular}{cc}
\includegraphics[width=0.4\linewidth, height=0.283\linewidth]{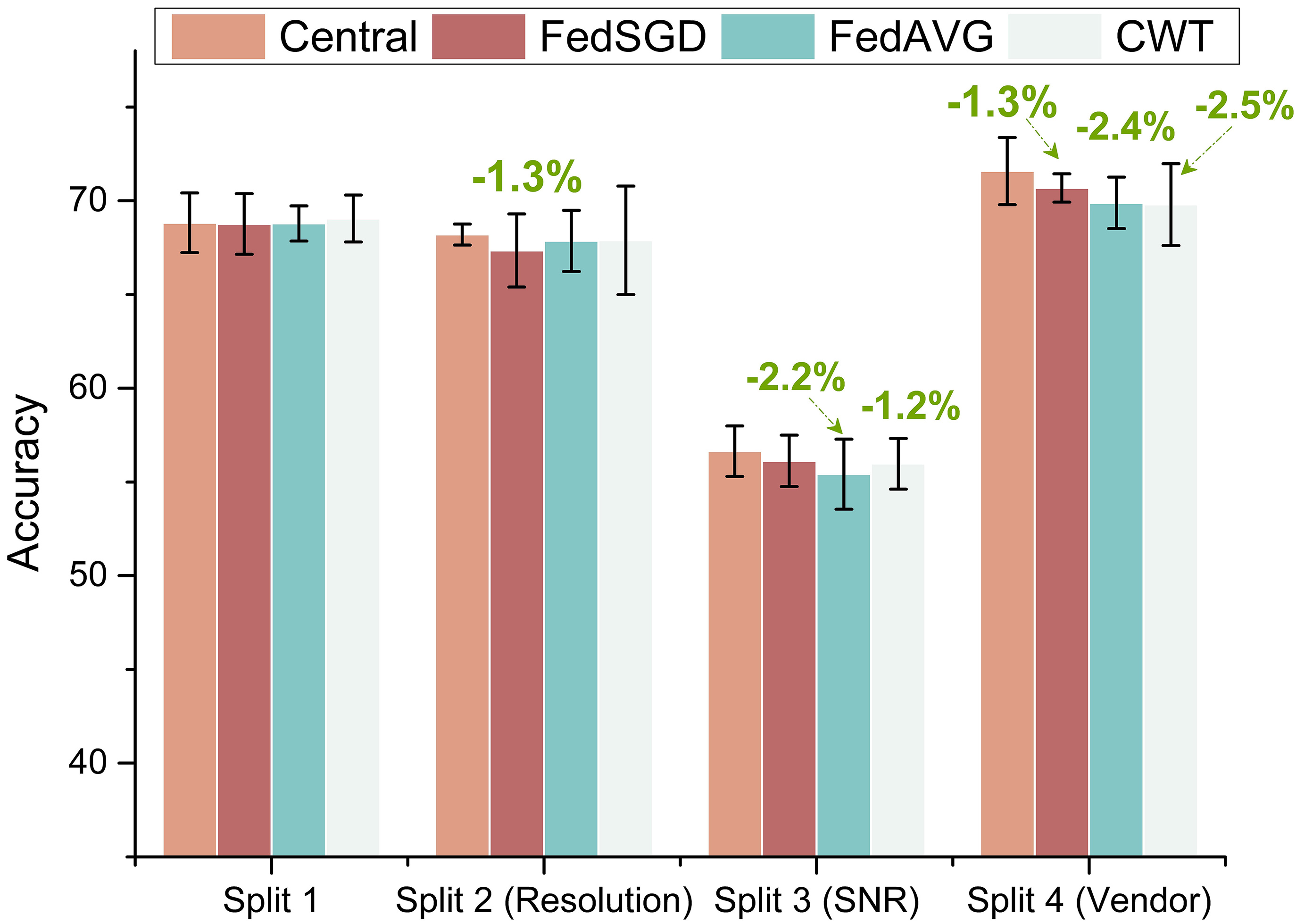}& %Im_var_ADNI  ADNI_im_var_1
  \includegraphics[width=0.37\linewidth, height=0.283\linewidth]{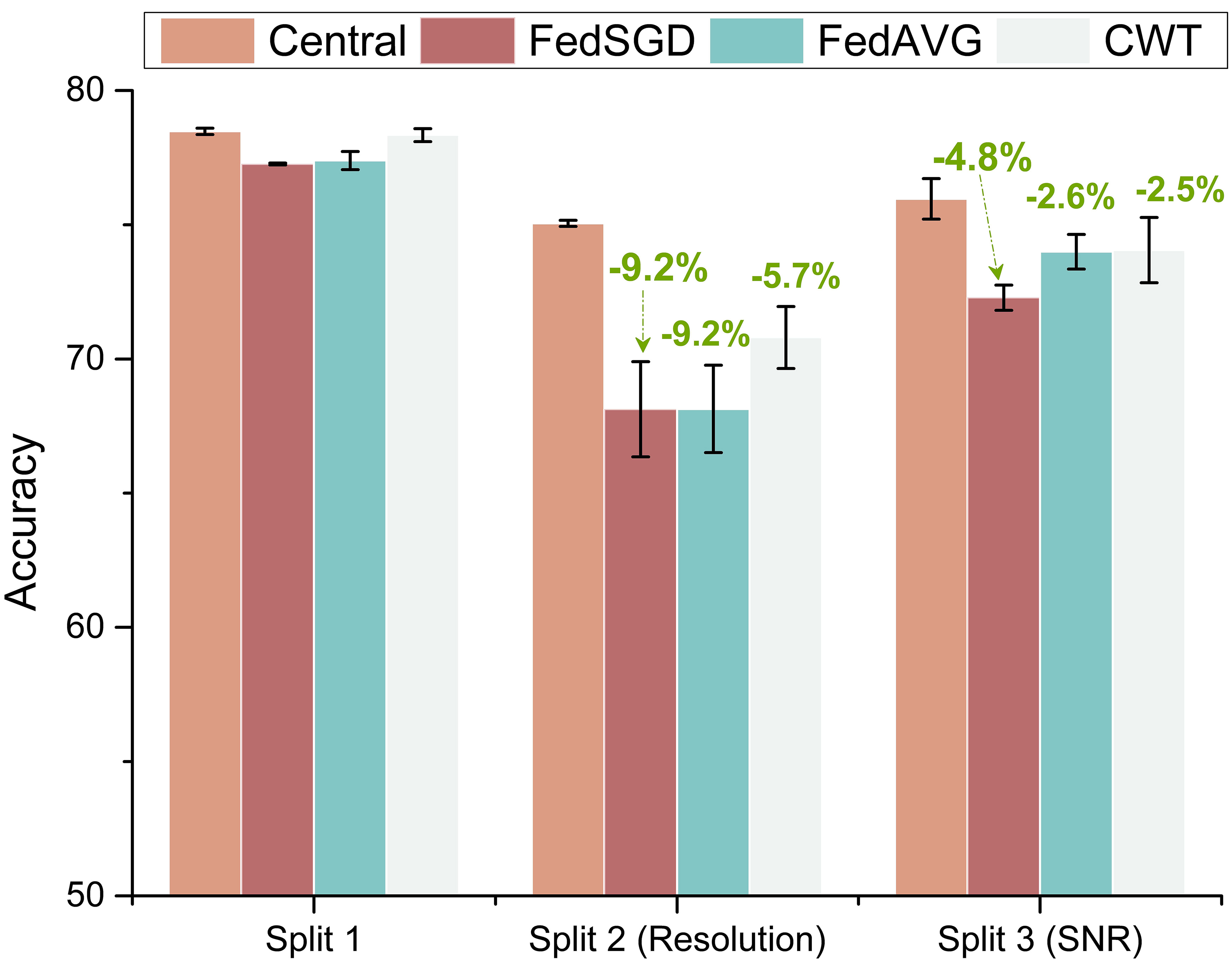}\\ %Im_var_Retina
     ADNI & Retina  \\
		\end{tabular}
	\end{center}
	\caption{Test accuracy on data partitions with imaging acquisition skew. The performance decrease rate compared to the baseline centrally hosted is also shown. }
	\label{fig:performance_im_var}
\end{figure*}

%Fig.~\ref{fig:performance_im_var} illustrates the test accuracy on data partitions with imaging acquisition skew. %Experimental results indicate that federated learning methods are more vulnerable to imaging acquisition skewness when its baseline centrally hosted is vulnerable to the image acquisition skew.
%The federated learning methods are more likely to retain the model accuracy in imaging acquisition skew when the baseline centrally hosted is robust to the imaging acquisition skew.
%For example, when centrally hosted shows comparable performance of split 2 and split 4 to the homogenous split 1 on ADNI dataset, the corresponding performance of federated learning methods on split 2 and split 4 is also acceptable, with performance drop rate less than 2.6\%.
%The federated learning methods are
%Once the baseline method is vulnerable to image acquisition variations,  On the other hand, large performance drop rates are detected on Retina dataset, where subtle variations of input image in Retina dataset could incur totally different outcome prediction.
Experimental results in Fig.~\ref{fig:performance_im_var} indicate that, in addition to the label distribution skew, the imaging acquisition skew is also a critical hurdle that prevents the deployment of federated learning in real applications. Strategies such as super-resolution, image denoising and histogram matching may be applied to deal with the imaging acquisition skew, which will remains to be undertaken in our future work. %The optimization methods for federated learning method which is invariable to feature distribution is highly required.

%Here we study the influence of feature distribution skew with several simple cases. In real application, the feature distribution across institutions maybe more complex and diverged.

%% catastrophic forgetting

\section{Conclusion}
In this paper, we conduct extensive experiments to study the impact of data heterogeneity on federated learning methods. Our study covers three widely used federated learning methods, a taxonomy of data heterogeneity regimes, data heterogeneity with different degrees of skewness.
We show that the federated learning methods in our study are vulnerable to data partitions with a high degree of skew. We then present several optimization strategies to overcome the performance loss from data heterogeneity.

Extensive experiments demonstrate that: 1) the proposed weighted average for FedSGD can recover performance loss from introducing quantity skew; 2) weighted loss helps mitigate the performance loss from introducing label distribution skew; 3) averaging the mean and variance of BN across institutions in FedAVG training is an attractive alternative to mitigate skew-induced performance loss of BN.
We anticipate that our detailed analysis provided herein will provide guidance for the deployment of federated learning in real clinical applications, and that our findings will provide useful hints towards the construction of better federated learning methods.

%Our future work will focus on further optimizing federated learning in data heterogeneity settings, and the study on data partitions with mixture of skew types.

% We also observe that federated learning methods are more vulnerable to image acquisition skewness when the study task is vulnerable to the feature distributions skew.

%We hope that the findings and insights in this
%paper, as well as our open source code, will spur further research into the fundamental and important problem of non-IID data in decentralized learning. employed on real applications with data collected from various institutions.

%These findings serve to address the challenges of real-world distributed
%deep learning tasks with medical image data, bringing federated learning
%closer to clinical implementation

%a real case scenario, the data across institutions will contain a mixture of skew types, which makes the problem even more challenging.
\clearpage

\section*{Supplementary Material}
\begin{figure}[h]
\centering
\includegraphics[width=0.95\linewidth]{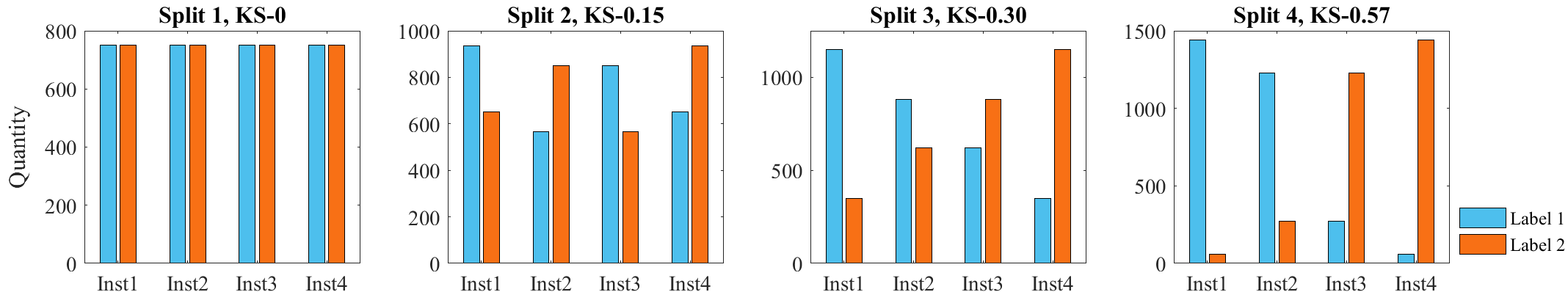}
\caption{Data partitions on Retina dataset with label distribution skew. Large KS indicates higher degree of label distribution skew.}
	\label{fig:Data_statis_label_var_Retina}
\end{figure}

We detail our simulated data partitions in this supplementary file. Table ~\ref{table:size_partition} shows the simulated data partitions with quantity skew.
Fig.~\ref{fig:Data_statis_label_var_ADNI} and Fig.~\ref{fig:Data_statis_label_var_Retina} show the simulated data partitions with label distribution skew on ADNI and Retina dataset, respectively. Fig.~\ref{fig:scanner_vendors} shows an example of PET cases imaged with different scanner vendors in ADNI dataset. These PET images differ in either resolution, contrast, or intensity distributions.

\begin{table}[h]
%\renewcommand{\tabcolsep}{8pt}
%\footnotesize
  \centering
  \caption{Data partitions with quantity skew. The number of training samples in each institution is shown. STD is sample standard deviation of the training sample size across institutions.}
  \label{table:size_partition}
  \subfloat[Partitions on ADNI]{%
    \hspace{.5cm}%
    \begin{tabular}{|c|c|c|c|c|c|}
        \hline
        Splits       & Inst1 & Inst2  & Inst3  & Inst4 & STD \\
        \hline
        \hline
        Split 1     & 474       &  474 &  474 &  474 & {0}  \\
        Split 2     & 299       &  317 &  385 &  895  & {283.1} \\
        Split 3     & 113       &  211 &  579 &  993  & {399.9} \\
        Split 4     & 66        & 111  &  282 &  1437 & {648.7} \\

        \hline
    \end{tabular}%
    \hspace{.5cm}%
  }
  \subfloat[Partitions on Retina]{%
    \hspace{.5cm}%
    \begin{tabular}{|c|c|c|c|c|c|}
        \hline
        Splits      & Inst1 & Inst2  & Inst2  & Inst2 & STD \\
        \hline
        \hline
        Split 1     & 1500       &  1500 &  1500 &  1500 & {0}  \\
        Split 2     & 750       &  960 &  1800& 2490 & {800.8} \\
        Split 3     &315	& 850	& 1750	& 3085	& {1211} \\
        Split 4     &208	& 350	& 889	& 4553	& {2056} \\
        \hline
    \end{tabular}%
   % \hspace{.5cm}%
  }
\end{table}

\begin{figure}[h]
\centering
\includegraphics[width=0.95\linewidth]{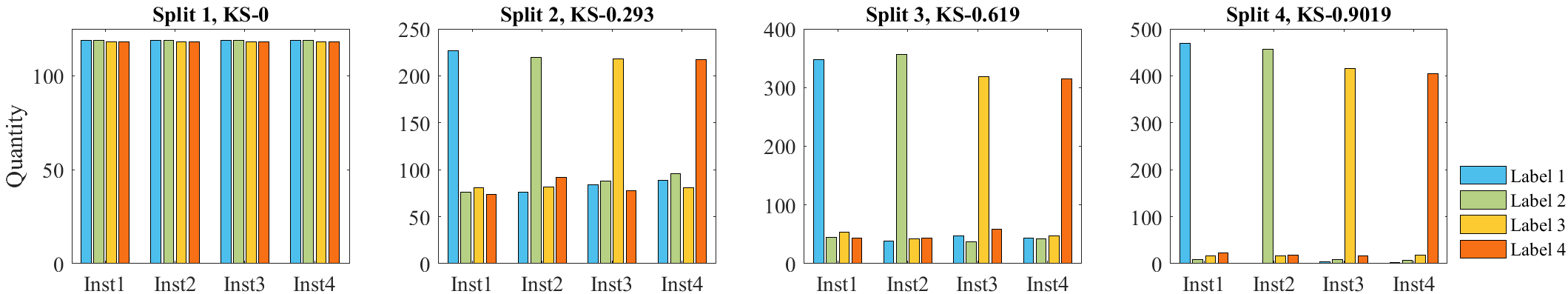}
\caption{Data partitions on ADNI dataset with label distribution skew. Large Kolmogorov-Smirnov (KS) indicates higher degree of label distribution skew.}
	\label{fig:Data_statis_label_var_ADNI}
\end{figure}

\begin{figure*}[h]
\scriptsize
	\begin{center}
		\begin{tabular}{ccccc}
  \includegraphics[width=0.18\linewidth]{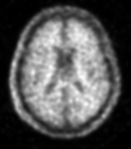}&
\includegraphics[width=0.18\linewidth]{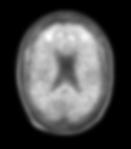}&
\includegraphics[width=0.18\linewidth]{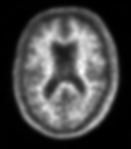}&
  \includegraphics[width=0.18\linewidth]{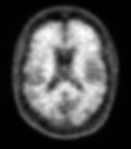}&
   \includegraphics[width=0.18\linewidth]{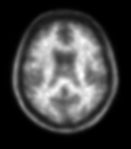} \\
   Philips & Siemens & CPS & GEMS & MiE\\
		\end{tabular}
	\end{center}
	\caption{PET cases imaged with different scanner vendors in ADNI dataset. }
	\label{fig:scanner_vendors}
\end{figure*}

\end{document}